\documentclass[11pt, a4paper, onecolumn, copyright, gdm]{google}

\usepackage[authoryear, sort&compress, round]{natbib}
\usepackage{multirow}

\uselogo{} 

\title{T5Gemma 2: Seeing, Reading, and Understanding Longer}

\renewcommand{\today}{}

\author[*]{Biao Zhang}
\author[*]{Paul Suganthan}
\author[*]{Gaël Liu}
\author[*]{Ilya Philippov}
\author[ \hspace{-0.3em}]{Sahil Dua}
\author[ \hspace{-0.3em}]{Ben Hora}
\author[ \hspace{-0.3em}]{Kat Black}
\author[ \hspace{-0.3em}]{Gus Martins}
\author[ \hspace{-0.3em}]{Omar Sanseviero}
\author[ \hspace{-0.3em}]{Shreya Pathak}
\author[ \hspace{-0.3em}]{Cassidy Hardin}
\author[ \hspace{-0.3em}]{Francesco Visin}
\author[ \hspace{-0.3em}]{Jiageng Zhang}
\author[ \hspace{-0.3em}]{Kathleen Kenealy}
\author[ \hspace{-0.3em}]{Qin Yin}
\author[ \hspace{-0.3em}]{Xiaodan Song}
\author[ \hspace{-0.3em}]{Olivier Lacombe}
\author[ \hspace{-0.3em}]{Armand Joulin}
\author[ \hspace{-0.3em}]{Tris Warkentin}
\author[$\dagger$]{Adam Roberts}

\correspondingauthor{biaojiaxing@google.com. \textnormal{Note, $^*$ Core contributors. $^\dagger$ Project Advisor.}}

\affil{\thepa{}{}}

\begin{abstract}
We introduce T5Gemma 2, the next generation of the T5Gemma family of lightweight open encoder-decoder models, featuring strong multilingual, multimodal and long-context capabilities. T5Gemma 2 follows the adaptation recipe (via UL2) in T5Gemma -- adapting a pretrained decoder-only model into an encoder-decoder model, and extends it from text-only regime to multimodal based on the Gemma 3 models. We further propose two methods to improve the efficiency: \textit{tied word embedding} that shares all embeddings across encoder and decoder, and \textit{merged attention} that unifies decoder self- and cross-attention into a single joint module. Experiments demonstrate the generality of the adaptation strategy over architectures and modalities as well as the unique strength of the encoder-decoder architecture on long context modeling. Similar to T5Gemma, T5Gemma 2 yields comparable or better pretraining performance and significantly improved post-training performance than its Gemma 3 counterpart. We release the pretrained models (270M-270M, 1B-1B and 4B-4B) to the community for future research.

\end{abstract}

\begin{document}

\maketitle

\begin{figure}[h]
    \small
    \centering
    \includegraphics[width=0.85\textwidth]{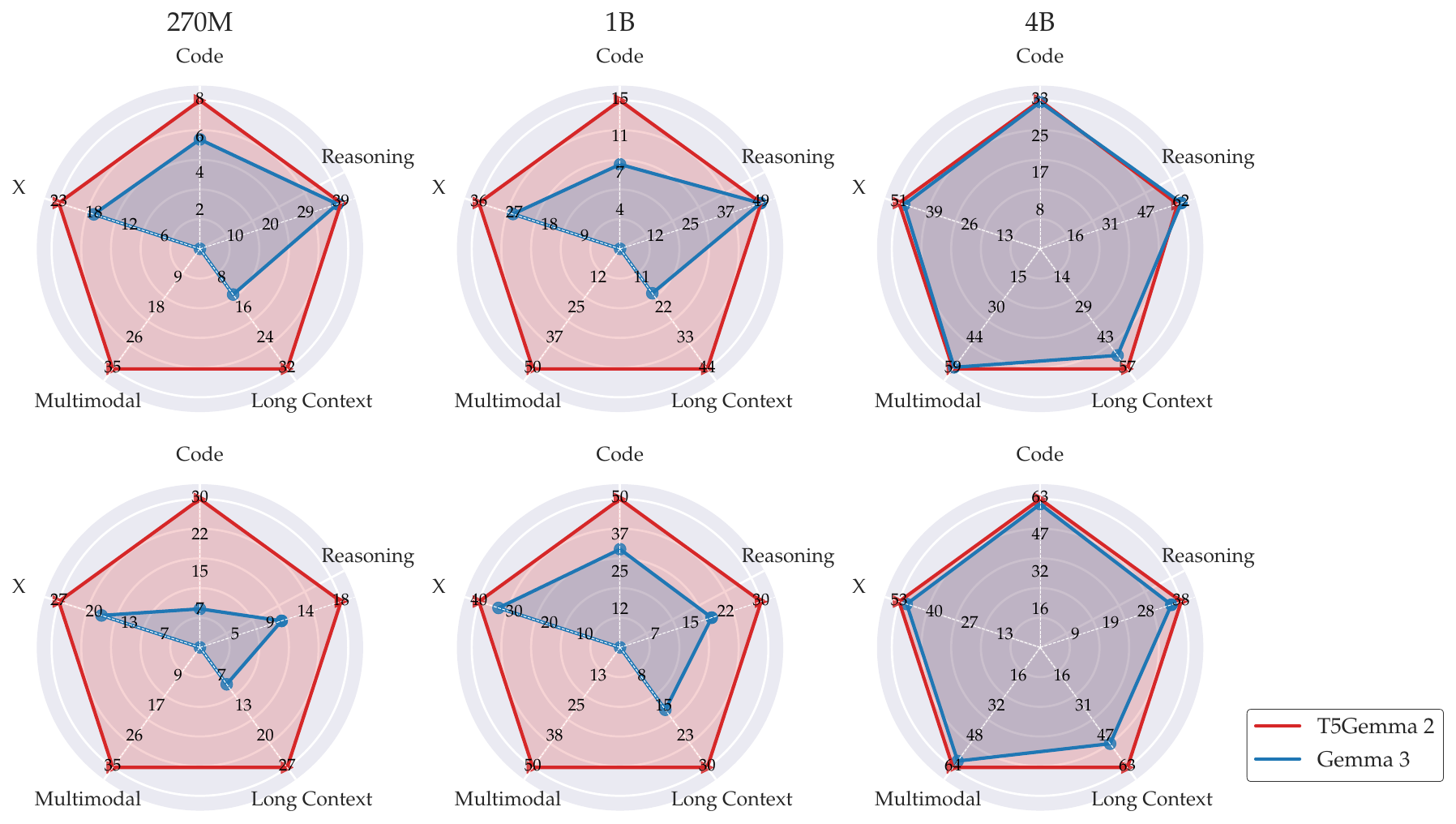}
    \caption{\label{fig:overall_performance} Summary of pretraining (top) and post-training (bottom) performance for Gemma 3 and T5Gemma 2 at 270M, 1B and 4B over five capabilities. \textit{270M} for T5Gemma 2 means a 270M encoder with a 270M decoder (the same applies to 1B and 4B). \textit{X}: multilingual. Note the post-training result for T5Gemma 2 is for illustration, where we only performed slight supervised finetuning without RL.}
\end{figure}

\section{Introduction}

The ability of jointly reading and perceiving under sufficient context has become essential in large language models (LLMs) for acquiring general world knowledge and advanced intelligence, as well as for many real-world applications, regardless of model architectures~\citep{comanici2025gemini,achiam2023gpt,claude3}.
In recent years, the encoder-decoder architecture has regained increasing interests in LLMs for its competitive scaling properties and flexible architectures~\citep{pmlr-v162-zhang22h,zhang2025redllm} and promising pre-/post-training performance~\citep{xue2021mt5,xue2022byt5,ao2022speecht5,tay2022ul2,pmlr-v162-wang22u,li2023openba,t5_paper,elfeki2025return}. Particularly, T5Gemma establishes modern, general-purpose encoder-decoder LLMs with non-trivial performance across benchmarks~\citep{zhang2025encoder}.
Still, the vast majority of these models (if not all) are \textit{blind}, operating exclusively on text-based data with limited context length -- a substantial gap compared to the advanced decoder-only LLMs~\citep{xu2025qwen3,team2025kimi,team2025gemma}.

\begin{figure}[t]
    \small
    \centering
    \includegraphics[width=0.85\textwidth]{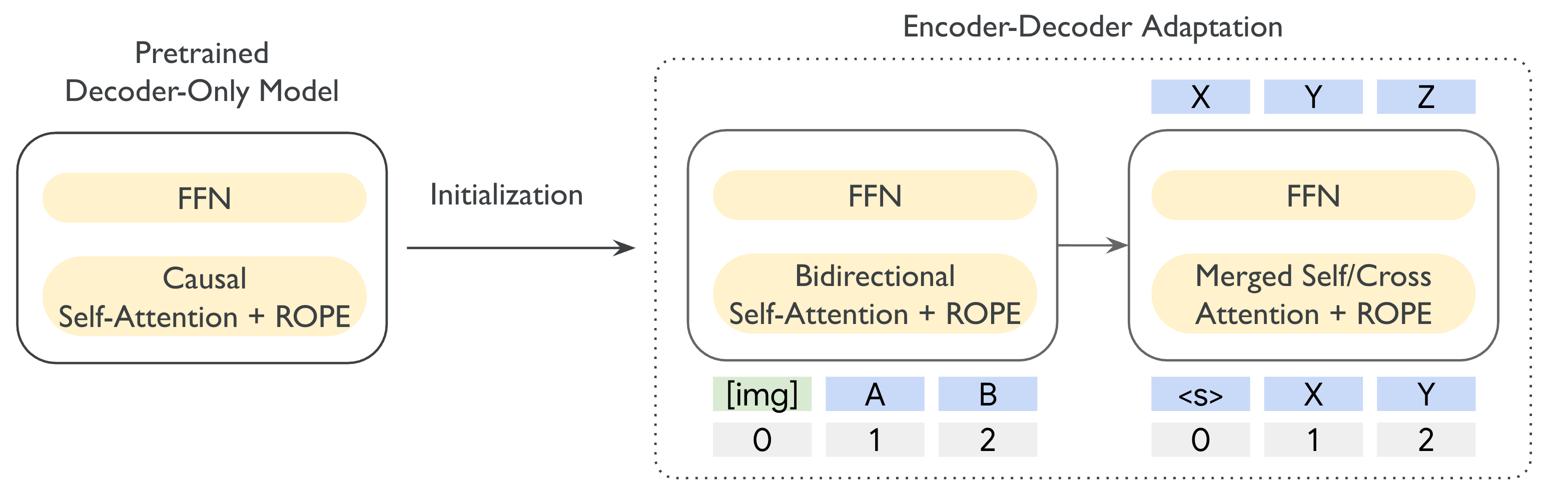}
    \caption{\label{fig:overview} Overview of T5Gemma 2. Encoder/decoder parameters are initialized from the pretrained decoder-only model, and then pretrained with UL2. We tie all word embeddings (in blue) and merge decoder self- and cross-attention sub-layers. Image is preprocessed by SigLIP into 256 embedding tokens and fed to the encoder for vision understanding (see \textit{[img]} for illustration).}
\end{figure}

In this paper, we fill this gap and present T5Gemma 2, a new family of lightweight open encoder-decoder LLMs with strong multilingual, multimodal and long-context capabilities. T5Gemma 2 follows the adaptation recipe from T5Gemma~\citep{zhang2025encoder}: initializing model parameters from a pretrained decoder-only checkpoint and then adapting them with the UL2 objective as in Figure \ref{fig:overview}. We further extend the recipe from the text-only realm to multimodal and long-context based on the powerful Gemma 3 models~\citep{team2025gemma}. For vision modeling, T5Gemma 2 reuses the same vision encoder from Gemma 3 and keeps it frozen; vision tokens are always fed to the encoder and all encoder tokens always have full visibility to each other in the self attention. For long-context modeling, we adopt the positional interpolation methods~\citep{chen2023extending,team2025gemma}. We also propose two strategies to save model parameters and improve the efficiency: 1) tying all word embeddings across encoder and decoder; and 2) merging the decoder self- and cross-attention into a single unified merged attention.

T5Gemma 2 features three model sizes: \textit{270M-270M}, \textit{1B-1B}, and \textit{4B-4B}. We pretrain each model on $\sim$2T tokens with an input/output sequence length up to 16K, and perform evaluation following Gemma 3. As in Figure \ref{fig:overall_performance}, the resulting T5Gemma 2 model shows competitive performance across different capabilities, matching and even surpassing its Gemma 3 counterpart in both pre- and post-training. Particularly, T5Gemma 2 270M-270M and 1B-1B yield encouraging multimodal performance even though their Gemma 3 base models are text-only, resonating with PaliGemma~\citep{steiner2024paligemma}. Besides, T5Gemma 2 delivers consistently improved long-context performance (up to 128K) despite of being pretrained on shorter sequences (only 16K), suggesting the special advantage of the encoder-decoder architecture on handling long context~\citep{zhang2025redllm}. We present the detailed results and also provide ablations justifying our architectural designs.

\begin{table*}[t]
\centering
\centering
\small

\begin{tabular}{lrl}
\toprule
Setting & Performance & \#Parameters \\
\midrule
Baseline & 47.8 & 4417M (1180M) \\
w/ Tied Embedding & 47.7 & 4417M (590M) \\
w/ Merged Attention & 47.5 & 4049M (1180M) \\
\midrule
w/ Cross Attention on Global Layers Only & 46.5 & 4233M (1180M) \\
\bottomrule
\end{tabular}
\caption{\label{tab:ablations} Architectural ablations for T5Gemma 2B-2B based on Gemma 2 2B. All models are pretrained on 400B tokens with PrefixLM+KD data. The performance is measured following the T5Gemma pretraining evaluation~\citep{zhang2025encoder}; \textit{\#Parameters} indicates the number of model (embedding) parameters.}
\end{table*}

\section{T5Gemma 2}

T5Gemma 2 is a Transformer-based encoder-decoder LLM~\citep{transformer}. Its basic building block follows Gemma 3: grouped-query attention~\citep{ainslie2023gqa} with QK-norm~\citep{dehghani2023scaling}, pre- and post-norm with RMSNorm~\citep{zhang2019root}, RoPE for positional encoding~\citep{su2024roformer}, and interleaved local and global attention layers with a ratio of 5:1. To improve long-context modeling, we set the RoPE base frequency to 10k and 1M for local and global attention layers, respectively~\citep{team2025gemma}. The 400M SigLIP encoder is adopted as the vision encoder~\citep{zhai2023sigmoid}, which transforms an image to 256 embedding tokens and is frozen during the training.

\paragraph{Tied Embedding}

T5Gemma uses separate word embeddings for the encoder and the decoder, which add a significant amount of model parameters particularly for small models.
In T5Gemma 2, we instead tie all word embeddings (encoder input embedding, decoder input embedding and decoder output/softmax embedding) following T5~\citep{t5_paper,press-wolf-2017-using}. Table \ref{tab:ablations} shows that tying embeddings leads to nearly no quality change but reduces the parameters by 10.5\%, suggesting the high redundancy of embedding parameters.

\paragraph{Merged Attention}

In the encoder-decoder architecture, cross-attention is often represented as a separated sub-layer in the decoder block, inserted in between the self-attention and feed-forward sub-layers. However, the functionality of the self- and cross-attention shares high similarity: gathering relevant information from the past. Inspired by previous studies~\citep{zhang2019improving,fu2023decoder,chowdhery2023palm}, we merge these two types of attentions into a single module with shared attention parameters.

Concretely, given the encoder output $\mathbf{H} \in \mathbb{R}^{n \times d}$ and the decoder self-attention input $\mathbf{X} \in \mathbb{R}^{m \times d}$, the merged attention operates as below:
\begin{align}
    \mathbf{Q} \in \mathbb{R}^{m\times d_h} = \mathbf{X} \mathbf{W}_q \\
    \mathbf{K} \in \mathbb{R}^{(m+n)\times d_h} = \left[\mathbf{X}; \mathbf{H}\right] \mathbf{W}_k \\
    \mathbf{V} \in \mathbb{R}^{(m+n)\times d_h} = \left[\mathbf{X}; \mathbf{H}\right] \mathbf{W}_v \\
    \mathbf{A} \in \mathbb{R}^{m\times d_h} = \text{SoftMax}\left( \frac{\mathbf{Q} \mathbf{K}^T}{\sqrt{d_h}} \odot \mathbf{M}\right) \mathbf{V} \\
    \mathbf{O} \in \mathbb{R}^{m\times d} = \mathbf{A}\mathbf{W}_o
\end{align}
where $n$/$m$ represents the encoder/decoder input length and $d$/$d_h$ is model/head dimension. For simplicity, we only describe the single head case. $\mathbf{W}_q, \mathbf{W}_k, \mathbf{W}_v \in \mathbb{R}^{d\times d_h}$ and $\mathbf{W}_o \in \mathbb{R}^{d_h\times d}$ are regular attention weight parameters. We concatenate the encoder output and decoder input $\left[\mathbf{X}; \mathbf{H}\right]$ such that both types of attention can be performed together. Note attention logits are also normalized jointly, similar to the decoder-only models. The masking $\mathbf{M}\in \mathbb{R}^{m\times (m+n)}$ handles the visibility to tokens from both encoder and decoder.

Merged attention narrows the architectural differences between the T5Gemma 2 decoder and the Gemma 3 decoder (see Figure \ref{fig:overview}), which eases the parameter initialization. Similarly, the encoder and decoder in T5Gemma 2 have roughly the same model parameters (see Table \ref{tab:model_params}). Ablations in Table \ref{tab:ablations} show that merged attention saves 6.5\% parameters and results in slight quality reduction\footnote{Note the parameter reduction (6.5\%) is based on the total parameters, i.e, both model and embedding parameters.}, $\sim$0.3 points on average, which we consider as an acceptable trade-off.

\paragraph{\textit{Rejected Ablation}: Cross-Attention on Global Layers Only}

By default, T5Gemma applies cross-attention to all decoder layers following the standard encoder-decoder architecture~\citep{transformer}. However, this adds non-ignorable computational cost particularly considering the autoregressive inference bottleneck and its global-attention structure. We thus explored a variant where we only apply it to decoder layers with global self-attention sub-layers, i.e. adding one cross-attention sub-layer every six decoder layers. Unfortunately, the experiments show a substantial quality drop by $\sim$1.3 points on average (see Table \ref{tab:ablations}). We consider this direction as reasonable but will need more efforts to retain the performance.

\begin{table*}[t]
\centering
\centering
\small

\begin{tabular}{lrrrr}
\toprule
\multirow{2}{*}{Model} & \multirow{2}{*}{Vision Encoder} & \multirow{2}{*}{Embedding} & \multicolumn{2}{c}{Non-Embedding} \\
\cmidrule(lr){4-5}
& & & Encoder & Decoder \\
\midrule
270M-270M & 417M & 168M & 100M & 100M  \\
1B-1B & 417M & 302M & 698M & 698M \\
4B-4B & 417M & 675M & 3209M & 3209M \\
\bottomrule
\end{tabular}
\caption{\label{tab:model_params} Number of parameters for T5Gemma 2 models.}
\end{table*}

\section{Setup}

\subsection{Pretraining}

\paragraph{Data}

Our pretraining data follows Gemma 3, which is a mixture of multilingual web documents, code, mathematical corpus and images~\citep{team2025gemma}. We preprocess the data with UL2~\citep{tay2022ul2} into $\leq$16K input sequences paired with $\leq$16K target outputs. Specifically, for text data, we apply the following five denoising tasks: $(\mu=3, r=0.15, n), (\mu=12, r=0.5, n), (\mu=32, r=0.15, n), (\mu=32, r=0.5, n)$ and $(\mu=\frac{3}{4}L, r=0.75, 1)$ with a mixing ratio of 1:1:1:1:4, where $\mu$ is the mean span length, $r$ is the corruption rate, $n$ is the number of corrupted spans, and $L$ is the input sequence length. We refer the readers to~\cite{tay2022ul2} for more details. For vision data, we only use prefix language modeling: all input tokens until the end of the final image are used as prefix, and the remaining text tokens are used as targets. Note, distillation is not used. The final pretraining data includes $\sim$2T tokens.

\paragraph{Optimization}

We initialize T5Gemma 2 parameters from the corresponding Gemma 3 pretraining checkpoint. All models are trained with a batch size of 4.2M tokens, and with the standard cross-entropy loss. Learning rate follows cosine decay with a warmup step of 100. To stabilize the training, we apply global gradient clipping at 1.0 and weight decay. We perform a simple grid search to decide the optimal learning rate for each model. The final pretraining checkpoint is created by averaging over the last 5 checkpoints (saved with an interval of 10K steps)~\cite{wortsman2022model}.

\begin{table*}[t]
\centering
\centering
\small

\begin{tabular}{llr}
\toprule
Model & {Setting} & Pretraining Performance \\
\midrule
\multirow{3}{*}{T5Gemma 2 270M-270M}
& PrefixLM + KD & 30.1 \\
& UL2 + KD & 30.5 \\
& UL2 & 30.6 \\
\midrule
\multirow{3}{*}{T5Gemma 2 1B-1B}
& PrefixLM + KD & 38.8 \\
& UL2 + KD & 40.2 \\
& UL2 & 39.8 \\
\bottomrule
\end{tabular}
\caption{\label{tab:data_ablations} Data ablations for T5Gemma 2. Note the pretraining performance is measured following T5Gemma~\citep{zhang2025encoder}, similar to Table \ref{tab:ablations}. UL2 and UL2+KD deliver comparable performance, which generally surpasses PrefixLM+KD.}
\end{table*}

\subsection{Post-training}

We also perform slight instruction tuning to showcase the strengths of encoder-decoder LLMs on downstream finetuning. Different from Gemma 3 post-training which includes distillation from stronger teacher and RL finetuning~\citep{team2025gemma}, we only apply distillation learning and train models with much less compute. Note the post-training performance in this paper should be considered as the lowerbound.

\subsection{Evaluation}

We evaluate the models following Gemma 3~\citep{team2025gemma}. Benchmarks include:
\begin{description}
    \item[Reasoning and factuality:] 
    HellaSwag~\citep{zellers2019hellaswag}, BoolQ~\citep{clark2019boolq}, PIQA~\citep{bisk2020piqa}, SIQA~\citep{sap2019socialiqa}, TriviaQA~\citep{2017arXivtriviaqa}, Natural Questions~\citep{kwiatkowski-etal-2019-natural}, ARC-C and ARC-E~\citep{allenai:arc}, WinoGrande~\citep{sakaguchi2021winogrande}, BBH~\citep{suzgun2022challenging}, DROP~\citep{dua2019drop}, and BIG-Bench Extra Hard~\citep{kazemi2025big}.

    \item[Stem and code:]
    MMLU-Pro~\citep{wang2024mmlu}, AGIEval~\citep{zhong2023agieval}, MATH~\citep{hendrycks2021measuring}, GSM8K~\citep{cobbe2021gsm8k}, GPQA~\citep{rein2023gpqa}, MBPP~\citep{austin2021program}, and HumanEval~\citep{chen2021evaluating}.

    \item[Multilingual:]
    MGSM~\citep{shi2022language}, Global-MMLU-Lite~\citep{singh2024global}, WMT24++~\citep{deutsch-etal-2025-wmt24}, FLoRes~\citep{goyal2022flores}, and XQuAD~\citep{artetxe2019cross}.

    \item[Multimodal:]
    COCO Caption~\citep{chen2015microsoft}, DocVQA~\citep{mathew2021docvqa}, InfographicVQA~\citep{mathew2022infographicvqa}, MMMU~\citep{yue2024mmmu}, TextVQA~\citep{singh2019towards}, RealWorldQA~\citep{realworldqa}, AI2D~\citep{kembhavi2016diagram}, ChartQA~\citep{masry2022chartqa}, VQA v2~\citep{goyal2017making}, TallyQA~\citep{acharya2019tallyqa}, and SpatialSense VQA~\citep{yang2019spatialsense}.

    \item[Long Context:]
    RULER~\citep{hsieh2024ruler} and MRCR~\citep{vodrahalli2024michelangelo}.
\end{description}

\section{Results}

\begin{table*}[h!]
\centering
\centering
\small

\begin{tabular}{llrrrrrrrr}
\toprule
& \multirow{2}{*}{Benchmark} & \multicolumn{3}{c}{Gemma 3} & \multicolumn{2}{c}{T5Gemma} & \multicolumn{3}{c}{T5Gemma 2} \\
\cmidrule(lr){3-5} \cmidrule(lr){6-7} \cmidrule(lr){8-10}
& & 270M & 1B & 4B & 2B-2B & 9B-9B & 270M-270M & 1B-1B & 4B-4B \\
\midrule
\multirow{12}{*}{\rotatebox[origin=c]{90}{\textit Reasoning and Factuality}}
& HellaSwag & 39.4 & 60.4 & 74.9 & 74.8 & \textbf{80.9} & 41.1 & 62.9 & 77.4 \\
& BoolQ & 60.5 & 66.1 & 79.1 & 75.5 & \textbf{85.7} & 57.4 & 68.5 & 79.3 \\
& PIQA & 67.7 & 74.7 & 79.7 & 79.0 & \textbf{81.1} & 66.5 & 74.3 & 79.2 \\
& SocialIQA & 45.0 & 46.9 & 49.0 & 49.8 & \textbf{50.1} & 46.1 & 47.7 & 49.9 \\
& TriviaQA & 14.3 & 39.8 & 65.7 & 51.2 & \textbf{75.2} & 14.8 & 29.0 & 53.1 \\
& Natural Questions & 2.8 & 9.6 & 20.2 & 15.1 & \textbf{28.6} & 3.0 & 8.9 & 17.2 \\
& ARC-c & 28.2 & 39.4 & 56.3 & 52.4 & \textbf{65.6} & 27.9 & 41.0 & 56.1 \\
& ARC-e & 56.1 & 72.2 & 81.9 & 77.1 & \textbf{85.4} & 57.9 & 67.7 & 74.3 \\
& WinoGrande & 51.9 & 58.7 & 68.0 & 69.9 & \textbf{78.8} & 53.9 & 60.8 & 71.6 \\
& BIG-Bench Hard & 23.6 & 28.3 & 50.9 & 54.8 & \textbf{74.4} & 22.9 & 28.5 & 43.7 \\
& DROP & 30.9 & 42.4 & 60.3 & 61.3 & \textbf{75.6} & 39.4 & 51.2 & 66.7 \\
\vspace{1pt}
& \textit{Average} & \textit{38.2} & \textit{49.0} & \textit{62.4} & \textit{60.1} & \textit{\textbf{71.0}} & \textit{39.2} & \textit{49.1} & \textit{60.8} \\
\midrule
\multirow{8}{*}{\rotatebox[origin=c]{90}{\textit Stem and Code}}
& MMLU (Pro COT) & 7.7 & 9.8 & 28.9 & 30.7 & \textbf{47.9} & 10.6 & 16.1 & 33.2 \\
& AGIEval & 20.8 & 21.8 & 41.6 & 35.2 & \textbf{53.6} & 20.9 & 23.8 & 41.1 \\
& MATH & 1.1 & 1.5 & 24.6 & 22.1 & \textbf{39.6} & 1.5 & 4.5 & 21.2 \\
& GSM8K & 0.5 & 1.3 & 38.1 & 46.7 & \textbf{74.7} & 1.7 & 9.1 & 44.0 \\
& GPQA & 6.9 & 9.6 & 15.0 & 18.5 & \textbf{23.9} & 9.6 & 10.5 & 17.6 \\
& MBPP & 0.6 & 9.4 & 45.8 & 38.0 & \textbf{53.8} & 5.6 & 25.2 & 44.8 \\
& HumanEval & 2.4 & 6.1 & 36.0 & 26.8 & \textbf{41.5} & 4.3 & 15.2 & 30.5 \\
\vspace{1pt}
& \textit{Average} & \textit{5.7} & \textit{8.5} & \textit{32.8} & \textit{31.1} & \textit{\textbf{47.9}} & \textit{7.7} & \textit{14.9} & \textit{33.2} \\
\midrule
\multirow{6}{*}{\rotatebox[origin=c]{90}{\textit Multilingual}}
& MGSM & 1.6 & 1.9 & 35.3 & 35.2 & \textbf{65.6} & 1.8 & 8.9 & 42.1 \\
& Global-MMLU-Lite & 27.0 & 24.5 & 57.0 & 40.3 & \textbf{66.1} & 23.4 & 33.1 & 53.5 \\
& WMT24++ (ChrF) & 18.3 & 36.7 & 48.4 & 40.8 & \textbf{51.5} & 26.9 & 40.9 & 49.2 \\
& FloRes & 20.3 & 29.6 & 39.2 & 31.8 & \textbf{42.8} & 23.9 & 33.8 & 41.8 \\
& XQuAD (all) & 20.7 & 43.8 & 68.1 & 58.0 & \textbf{73.8} & 40.8 & 63.1 & 70.6 \\
\vspace{1pt}
& \textit{Average} & \textit{17.6} & \textit{27.3} & \textit{49.6} & \textit{41.2} & \textit{\textbf{59.9}} & \textit{23.4} & \textit{36.0} & \textit{51.4} \\
\midrule
\multirow{12}{*}{\rotatebox[origin=c]{90}{\textit Multimodal}}
& COCOcap & - & - & 101.8 & - & - & 69.6 & 86.0 & \textbf{105.4} \\
& DocVQA (val)$^*$ & - & - & 73.0 & - & - & 41.6 & 66.6 & \textbf{74.7} \\
& InfoVQA (val)$^*$ & - & - & 44.0 & - & - & 20.2 & 36.4 & \textbf{46.0} \\
& MMMU (pt) & - & - & 37.9 & - & - & 22.7 & 28.4 & \textbf{39.4} \\
& TextVQA (val) & - & - & \textbf{58.7} & - & - & 34.1 & 53.1 & 58.4 \\
& RealWorldQA & - & - & 43.5 & - & - & 27.2 & 42.4 & \textbf{46.1} \\
& AI2D & - & - & \textbf{63.1} & - & - & 26.5 & 44.8 & 61.6 \\
& ChartQA & - & - & 63.2 & - & - & 29.2 & 50.2 & \textbf{66.0} \\
& VQAv2 & - & - & \textbf{63.8} & - & - & 38.8 & 57.8 & 62.7 \\
& TallyQA & - & - & \textbf{42.8} & - & - & 26.4 & 32.2 & 39.6 \\
& SpatialSense VQA & - & - & 51.3 & - & - & 50.4 & 50.2 & \textbf{51.7} \\
\vspace{1pt}
& \textit{Average} & - & - & \textit{58.5} & - & - & \textit{35.1} & \textit{49.8} & \textit{\textbf{59.2}} \\
\midrule
\multirow{5}{*}{\rotatebox[origin=c]{90}{\textit Long Context}}
& Ruler 32K & 21.3 & 23.3 & 66.8 & 0.2 & 0.0 & 57.3 & 69.2 & \textbf{81.7} \\
& Ruler 128K & 4.4 & 7.0 & 51.7 & 0.5 & 0.5 & 25.5 & 35.1 & \textbf{57.6} \\
& MRCR 32K & 14.6 & 20.8 & 46.5 & 35.3 & 46.0 & 23.4 & 38.6 & \textbf{49.4} \\
& MRCR 128K & 7.8 & 13.8 & 37.8 & 15.6 & 22.3 & 20.4 & 32.5 & \textbf{39.8} \\
\vspace{1pt}
& \textit{Average} & \textit{12.0} & \textit{16.2} & \textit{50.7} & \textit{12.9} & \textit{17.2} & \textit{31.7} & \textit{43.8} & \textit{\textbf{57.1}} \\
\bottomrule
\end{tabular}
\caption{\label{tab:pt_result} Detailed pretraining results for Gemma 3, T5Gemma, and T5Gemma 2. Note Gemma 3 270M\&1B and T5Gemma 2B-2B\&9B-9B are text-only models. $^*$: approximate results which can't be compared across papers. In general, T5Gemma 2 shows strong multi-modal and long-context performance. Best results are highlighted in \textbf{bold}.}
\end{table*}

\begin{table*}[t]
\centering
\centering
\small

\begin{tabular}{llrrrrrrrr}
\toprule
& \multirow{2}{*}{Benchmark} & \multicolumn{3}{c}{Gemma 3} & \multicolumn{2}{c}{T5Gemma} & \multicolumn{3}{c}{T5Gemma 2} \\
\cmidrule(lr){3-5} \cmidrule(lr){6-7} \cmidrule(lr){8-10}
& & 270M & 1B & 4B & 2B-2B & 9B-9B & 270M-270M & 1B-1B & 4B-4B \\
\midrule
\multirow{4}{*}{\rotatebox[origin=c]{90}{\textit Reasoning}}
& GPQA Diamond & 19.7 & 14.1 & 24.7 & 20.7 & \textbf{33.3} & 16.2 & 23.2 & 27.8 \\
& BIG-Bench Hard & 10.0 & 36.3 & 71.7 & 58.2 & \textbf{80.4} & 34.6 & 59.2 & 75.0 \\
& BIG-Bench Extra Hard & 1.7 & 7.2 & 8.7 & 9.1 & \textbf{10.6} & 3.5 & 6.3 & 10.4 \\
\vspace{1pt}
& \textit{Average} & \textit{10.5} & \textit{19.2} & \textit{35.1} & \textit{29.3} & \textit{\textbf{41.5}} & \textit{18.1} & \textit{29.6} & \textit{37.7} \\
\midrule
\multirow{7}{*}{\rotatebox[origin=c]{90}{\textit Stem and Code}}
& MMLU (Pro) & 1.3 & 13.2 & 40.9 & 32.3 & \textbf{54.5} & 10.5 & 25.9 & 44.4 \\
& HiddenMath & 0.4 & 11.4 & \textbf{35.1} & 4.4 & 13.3 & 3.3 & 17.3 & 28.8 \\
& MBPP & 11.4 & 36.0 & 62.6 & 43.8 & 64.4 & 32.0 & 54.8 & \textbf{66.8} \\
& HumanEval & 14.6 & 43.9 & 70.7 & 46.3 & 74.4 & 42.1 & 61.6 & \textbf{76.8} \\
& Natural2Code & 16.9 & 58.0 & 72.0 & 56.0 & \textbf{75.4} & 54.6 & 68.1 & 72.9 \\
& GSM8K & 1.9 & 35.9 & 84.0 & 69.7 & \textbf{89.9} & 36.5 & 72.1 & 88.6 \\
\vspace{1pt}
& \textit{Average} & \textit{7.8} & \textit{33.1} & \textit{60.9} & \textit{42.1} & \textit{62.0} & \textit{29.8} & \textit{50.0} & \textit{\textbf{63.1}} \\
\midrule
\multirow{3}{*}{\rotatebox[origin=c]{90}{\tiny \textit Multilingual}}
& Global MMLU Lite & 19.7 & 34.9 & 53.4 & 50.4 & \textbf{67.3} & 28.6 & 42.3 & 59.6 \\
& WMT24++ & 17.5 & 34.7 & 47.0 & 39.1 & \textbf{48.0} & 24.6 & 38.5 & 46.5 \\
\vspace{1pt}
& \textit{Average} & \textit{18.6} & \textit{34.8} & \textit{50.2} & \textit{44.8} & \textit{\textbf{57.7}} & \textit{26.6} & \textit{40.4} & \textit{53.0} \\
\midrule
\multirow{9}{*}{\rotatebox[origin=c]{90}{\textit Multimodal}}
& MMMU (val) & - & - & 47.3 & - & - & 30.8 & 37.8 & \textbf{47.8} \\
& DocVQA$^*$ & - & - & 74.1 & - & - & 42.2 & 63.2 & \textbf{75.2} \\
& InfoVQA$^*$ & - & - & 40.0 & - & - & 19.9 & 33.4 & \textbf{43.5} \\
& TextVQA & - & - & 57.5 & - & - & 35.0 & 28.5 & \textbf{59.0} \\
& AI2D & - & - & 75.0 & - & - & 48.8 & 69.6 & \textbf{78.3} \\
& ChartQA & - & - & 69.0 & - & - & 36.4 & 56.8 & \textbf{75.0} \\
& VQAv2 (val) & - & - & 59.3 & - & - & 29.2 & 61.2 & \textbf{67.3} \\
\vspace{1pt}
& \textit{Average} & - & - & \textit{60.3} & - & - & \textit{34.6} & \textit{50.1} & \textit{\textbf{63.7}} \\
\midrule
\multirow{5}{*}{\rotatebox[origin=c]{90}{\textit Long Context}}
& Ruler 32K & 16.2 & 31.2 & 61.3 & 0.1 & 0.7 & 52.9 & 47.5 & \textbf{83.1} \\
& Ruler 128K & 0.0 & 2.5 & \textbf{46.9} & 0.2 & 0.2 & 11.2 & 6.4 & 39.5 \\
& MRCR 32K & 10.1 & 14.7 & 50.3 & 53.6 & 56.7 & 20.9 & 34.8 & \textbf{69.8} \\
& MRCR 128K & 6.6 & 14.0 & 41.9 & 24.3 & 30.3 & 21.9 & 31.5 & \textbf{57.7} \\
\vspace{1pt}
& \textit{Average} & \textit{8.2} & \textit{15.6} & \textit{50.1} & \textit{19.6} & \textit{21.9} & \textit{26.8} & \textit{30.1} & \textit{\textbf{62.5}} \\
\bottomrule
\end{tabular}
\caption{\label{tab:it_result} Detailed post-training results for Gemma 3, T5Gemma, and T5Gemma 2. T5Gemma 2 outperforms Gemma 3 on most capabilities despite its lightweight post-training.}
\end{table*}

\paragraph{PrefixLM+KD, UL2, vs. UL2+KD: Do distillation and training objective matter?}

T5Gemma ablates the effect of different pretraining data (PrefixLM+KD vs. UL2), showing mixed results. In T5Gemma 2, we further examine these options and also compare to UL2+KD, where we use teacher logits for real target tokens and one-hot logits for special masking tokens.

Table \ref{tab:data_ablations} shows that PrefixLM+KD generally performs the worst while UL2(+KD) is consistently better for models $\leq$ 1B-1B. UL2+KD performs slightly better at 1B-1B than UL2 by $\sim$0.4 points on average. With T5Gemma results, we argue that the effect of distillation highly depends on the teacher and student modeling capacity~\citep{zhou2019understanding}. We decided to drop the distillation due to its expensive data loading overhead and simply use UL2 for T5Gemma 2.

\paragraph{Text-only LLMs can be adapted into strong multimodal and long-context encoder-decoder models.}

While Gemma 3 270M and 1B are text-only and context limited, Table \ref{tab:pt_result} shows that our adaptation recipe successfully adapts them into multimodal and long-context with non-trivial performance, resonating with previous findings~\citep{steiner2024paligemma,chen2023extending}. For example, T5Gemma 2 1B-1B yields an average multimodal and long-context result of 49.8 and 43.8, lagging behind Gemma 3 4B by only $8.7$ and $6.9$ points, respectively, despite being much smaller. We ascribe this to the special architecture of encoder-decoder models, where the encoder parameters are exclusively used for input/vision understanding with bidirectional attention, and the cross-attention allows for attending to high-level representations of the input.

\paragraph{T5Gemma 2 achieves competitive pretraining performance and improved post-training performance than Gemma 3.}

Overall, T5Gemma 2 270M-270M and 1B-1B substantially outperform Gemma 3 270M and 1B after pretraining across benchmarks, respectively. It performs on par with or slightly better than Gemma 3 at 4B-4B scale, as shown in Table \ref{tab:pt_result}. After post-training, T5Gemma 2 generally surpasses Gemma 3 despite its lightweight finetuning, as shown in Table \ref{tab:it_result}, echoing with previous findings~\citep{zhang2025encoder,zhang2025redllm,pmlr-v162-wang22u}. Note the post-training result for T5Gemma 2 is for illustration only, and we believe it could be significantly enhanced with comprehensive RL learning.

We note that T5Gemma 2 shows consistently better long-context and multi-modal performance than Gemma 3 and T5Gemma. This demonstrates 1) the adaptation recipe from T5Gemma generalizes across modalities, and 2) the unique adaptability of encoder-decoder models. We hope these insights can inspire further exploration on the encoder-decoder architecture for general-purpose language modeling.

\section{Conclusion}

We have presented T5Gemma 2, the new collection of vision-language encoder-decoder foundation model. T5Gemma 2 was built by adapting the pretrained decoder-only Gemma 3 models into encoder-decoder on $\sim$2T UL2 tokens. We ablated several architecture designs and integrated two proposal to save model parameters: \textit{tied embeddings} across encoder and decoder, and \textit{merged attention} unifying decoder self- and cross-attention sub-layers. The resulting decoder architecture resembles the encoder architecture, facilitating the adaptation from decoder-only models.

T5Gemma 2 accepts text and/or image as inputs to the encoder, and generates response text from the decoder. We evaluated the models across a range of benchmarks, covering five capabilities: reasoning and factuality, stem and coding, multilingual, multimodal and long-context. In general, T5Gemma 2 shows competitive pretraining performance than Gemma 3 and improved post-training performance across capabilities.

Especially, T5Gemma 2 shows strong multimodal and long-context performance, thanks to its encoder-decoder architecture. Unlike decoder-only LLMs, T5Gemma 2 has a dedicated set of encoder parameters for input/vision or prompt understanding. Its cross-attention over the encoder outputs also endows it with better ability on retrieving relevant information from the inputs.

Beyond standalone usage, T5Gemma 2 serves as a robust foundation for training high-quality downstream embedding models. For example, EmbeddingGemma \citep{vera2025embeddinggemma} leverages T5Gemma 2 checkpoints to achieve state-of-the-art performance on text retrieval benchmarks.

We released all three pretrained checkpoints (270M-270M, 1B-1B and 4B-4B) to facilitate the evaluation, adaptation and research by the community. Note, to the best of our knowledge, T5Gemma 2 presents itself as the first capable long-context encoder-decoder LLMs (up to 128K) in the community. By offering novel insights into encoder-decoder LLMs, we hope this work to be a catalyst for future innovation, ultimately benefiting the development of more sophisticated and powerful LLMs.

\section*{Acknowledgments}
We'd like to thank Lechao Xiao for his insightful comments.
Our work is made possible by the dedication and efforts of numerous teams at Google. We would like to acknowledge the support from the following teams: DevX, Gemini Infrastructure, Gemini Safety, Gemma, Google Cloud, Google Research Responsible AI, Kaggle, and Gemini Encoder-heavy.

\bibliography{main}

@article{zhang2025encoder,
  title={Encoder-Decoder Gemma: Improving the Quality-Efficiency Trade-Off via Adaptation},
  author={Zhang, Biao and Moiseev, Fedor and Ainslie, Joshua and Suganthan, Paul and Ma, Min and Bhupatiraju, Surya and Lebron, Fede and Firat, Orhan and Joulin, Armand and Dong, Zhe},
  journal={arXiv preprint arXiv:2504.06225},
  year={2025}
}

@article{team2025gemma,
  title={Gemma 3 technical report},
  author={Team, Gemma and Kamath, Aishwarya and Ferret, Johan and Pathak, Shreya and Vieillard, Nino and Merhej, Ramona and Perrin, Sarah and Matejovicova, Tatiana and Ram{\'e}, Alexandre and Rivi{\`e}re, Morgane and others},
  journal={arXiv preprint arXiv:2503.19786},
  year={2025}
}

@article{tay2022ul2,
  title={Ul2: Unifying language learning paradigms},
  author={Tay, Yi and Dehghani, Mostafa and Tran, Vinh Q and Garcia, Xavier and Wei, Jason and Wang, Xuezhi and Chung, Hyung Won and Shakeri, Siamak and Bahri, Dara and Schuster, Tal and others},
  journal={arXiv preprint arXiv:2205.05131},
  year={2022}
}

@article{zhang2019root,
  title={Root mean square layer normalization},
  author={Zhang, Biao and Sennrich, Rico},
  journal={Advances in neural information processing systems},
  volume={32},
  year={2019}
}

@article{zhang2025redllm,
  title={Encoder-Decoder or Decoder-Only? Revisiting Encoder-Decoder Large Language Model},
  author={Zhang, Biao and Cheng, Yong and Shakeri, Siamak and Wang, Xinyi and Ma, Min and Firat, Orhan},
  journal={arXiv preprint arXiv:2510.26622},
  year={2025}
}

@article{comanici2025gemini,
  title={Gemini 2.5: Pushing the frontier with advanced reasoning, multimodality, long context, and next generation agentic capabilities},
  author={Comanici, Gheorghe and Bieber, Eric and Schaekermann, Mike and Pasupat, Ice and Sachdeva, Noveen and Dhillon, Inderjit and Blistein, Marcel and Ram, Ori and Zhang, Dan and Rosen, Evan and others},
  journal={arXiv preprint arXiv:2507.06261},
  year={2025}
}

@article{achiam2023gpt,
  title={Gpt-4 technical report},
  author={Achiam, Josh and Adler, Steven and Agarwal, Sandhini and Ahmad, Lama and Akkaya, Ilge and Aleman, Florencia Leoni and Almeida, Diogo and Altenschmidt, Janko and Altman, Sam and Anadkat, Shyamal and others},
  journal={arXiv preprint arXiv:2303.08774},
  year={2023}
}

@article{claude3,
  title={The Claude 3 Model Family: Opus, Sonnet, Haiku},
  author={Anthropic},
  url={https://www.anthropic.com/},
  journal={},
  year={2024},
}

@article{xu2025qwen3,
  title={Qwen3-omni technical report},
  author={Xu, Jin and Guo, Zhifang and Hu, Hangrui and Chu, Yunfei and Wang, Xiong and He, Jinzheng and Wang, Yuxuan and Shi, Xian and He, Ting and Zhu, Xinfa and others},
  journal={arXiv preprint arXiv:2509.17765},
  year={2025}
}

@article{team2025kimi,
  title={Kimi-vl technical report},
  author={Team, Kimi and Du, Angang and Yin, Bohong and Xing, Bowei and Qu, Bowen and Wang, Bowen and Chen, Cheng and Zhang, Chenlin and Du, Chenzhuang and Wei, Chu and others},
  journal={arXiv preprint arXiv:2504.07491},
  year={2025}
}

@InProceedings{pmlr-v162-zhang22h,
  title = 	 {Examining Scaling and Transfer of Language Model Architectures for Machine Translation},
  author =       {Zhang, Biao and Ghorbani, Behrooz and Bapna, Ankur and Cheng, Yong and Garcia, Xavier and Shen, Jonathan and Firat, Orhan},
  booktitle = 	 {Proceedings of the 39th International Conference on Machine Learning},
  pages = 	 {26176--26192},
  year = 	 {2022},
  editor = 	 {Chaudhuri, Kamalika and Jegelka, Stefanie and Song, Le and Szepesvari, Csaba and Niu, Gang and Sabato, Sivan},
  volume = 	 {162},
  series = 	 {Proceedings of Machine Learning Research},
  month = 	 {17--23 Jul},
  publisher =    {PMLR},
  url = 	 {https://proceedings.mlr.press/v162/zhang22h.html},
}

@article{t5_paper,
    author = {Raffel, Colin and Shazeer, Noam and Roberts, Adam and Lee, Katherine and Narang, Sharan and Matena, Michael and Zhou, Yanqi and Li, Wei and Liu, Peter J.},
    title = {Exploring the Limits of Transfer Learning with a Unified Text-to-Text Transformer},
    year = {2020},
    issue_date = {January 2020},
    publisher = {JMLR.org},
    volume = {21},
    number = {1},
    issn = {1532-4435},
    month = {jan},
    articleno = {140},
    numpages = {67},
}

@inproceedings{transformer,
 author = {Vaswani, Ashish and Shazeer, Noam and Parmar, Niki and Uszkoreit, Jakob and Jones, Llion and Gomez, Aidan N and Kaiser, \L ukasz and Polosukhin, Illia},
 booktitle = {Advances in Neural Information Processing Systems},
 editor = {I. Guyon and U. Von Luxburg and S. Bengio and H. Wallach and R. Fergus and S. Vishwanathan and R. Garnett},
 pages = {},
 publisher = {Curran Associates, Inc.},
 title = {Attention is All you Need},
 url = {https://proceedings.neurips.cc/paper_files/paper/2017/file/3f5ee243547dee91fbd053c1c4a845aa-Paper.pdf},
 volume = {30},
 year = {2017}
}

@InProceedings{pmlr-v162-wang22u,
  title = 	 {What Language Model Architecture and Pretraining Objective Works Best for Zero-Shot Generalization?},
  author =       {Wang, Thomas and Roberts, Adam and Hesslow, Daniel and Scao, Teven Le and Chung, Hyung Won and Beltagy, Iz and Launay, Julien and Raffel, Colin},
  booktitle = 	 {Proceedings of the 39th International Conference on Machine Learning},
  pages = 	 {22964--22984},
  year = 	 {2022},
  editor = 	 {Chaudhuri, Kamalika and Jegelka, Stefanie and Song, Le and Szepesvari, Csaba and Niu, Gang and Sabato, Sivan},
  volume = 	 {162},
  series = 	 {Proceedings of Machine Learning Research},
  month = 	 {17--23 Jul},
  publisher =    {PMLR},
  url = 	 {https://proceedings.mlr.press/v162/wang22u.html},
}

@article{li2023openba,
  title={OpenBA: An Open-sourced 15B Bilingual Asymmetric seq2seq Model Pre-trained from Scratch},
  author={Li, Juntao and Tang, Zecheng and Ding, Yuyang and Wang, Pinzheng and Guo, Pei and You, Wangjie and Qiao, Dan and Chen, Wenliang and Fu, Guohong and Zhu, Qiaoming and others},
  journal={arXiv preprint arXiv:2309.10706},
  year={2023}
}

@article{elfeki2025return,
  title={Return of the Encoder: Maximizing Parameter Efficiency for SLMs},
  author={Elfeki, Mohamed and Liu, Rui and Voegele, Chad},
  journal={arXiv preprint arXiv:2501.16273},
  year={2025}
}

@article{steiner2024paligemma,
  title={Paligemma 2: A family of versatile vlms for transfer},
  author={Steiner, Andreas and Pinto, Andr{\'e} Susano and Tschannen, Michael and Keysers, Daniel and Wang, Xiao and Bitton, Yonatan and Gritsenko, Alexey and Minderer, Matthias and Sherbondy, Anthony and Long, Shangbang and others},
  journal={arXiv preprint arXiv:2412.03555},
  year={2024}
}

@article{ainslie2023gqa,
  title={Gqa: Training generalized multi-query transformer models from multi-head checkpoints},
  author={Ainslie, Joshua and Lee-Thorp, James and De Jong, Michiel and Zemlyanskiy, Yury and Lebr{\'o}n, Federico and Sanghai, Sumit},
  journal={arXiv preprint arXiv:2305.13245},
  year={2023}
}

@inproceedings{zhai2023sigmoid,
  title={Sigmoid loss for language image pre-training},
  author={Zhai, Xiaohua and Mustafa, Basil and Kolesnikov, Alexander and Beyer, Lucas},
  booktitle={Proceedings of the IEEE/CVF international conference on computer vision},
  pages={11975--11986},
  year={2023}
}

@inproceedings{dehghani2023scaling,
  title={Scaling vision transformers to 22 billion parameters},
  author={Dehghani, Mostafa and Djolonga, Josip and Mustafa, Basil and Padlewski, Piotr and Heek, Jonathan and Gilmer, Justin and Steiner, Andreas Peter and Caron, Mathilde and Geirhos, Robert and Alabdulmohsin, Ibrahim and others},
  booktitle={International conference on machine learning},
  pages={7480--7512},
  year={2023},
  organization={PMLR}
}

@article{su2024roformer,
  title={Roformer: Enhanced transformer with rotary position embedding},
  author={Su, Jianlin and Ahmed, Murtadha and Lu, Yu and Pan, Shengfeng and Bo, Wen and Liu, Yunfeng},
  journal={Neurocomputing},
  volume={568},
  pages={127063},
  year={2024},
  publisher={Elsevier}
}

@article{zhang2019improving,
  title={Improving deep transformer with depth-scaled initialization and merged attention},
  author={Zhang, Biao and Titov, Ivan and Sennrich, Rico},
  journal={arXiv preprint arXiv:1908.11365},
  year={2019}
}

@article{fu2023decoder,
  title={Decoder-only or encoder-decoder? interpreting language model as a regularized encoder-decoder},
  author={Fu, Zihao and Lam, Wai and Yu, Qian and So, Anthony Man-Cho and Hu, Shengding and Liu, Zhiyuan and Collier, Nigel},
  journal={arXiv preprint arXiv:2304.04052},
  year={2023}
}

@article{chowdhery2023palm,
  title={Palm: Scaling language modeling with pathways},
  author={Chowdhery, Aakanksha and Narang, Sharan and Devlin, Jacob and Bosma, Maarten and Mishra, Gaurav and Roberts, Adam and Barham, Paul and Chung, Hyung Won and Sutton, Charles and Gehrmann, Sebastian and others},
  journal={Journal of Machine Learning Research},
  volume={24},
  number={240},
  pages={1--113},
  year={2023}
}

@inproceedings{
hendrycks2021measuring,
title={Measuring Massive Multitask Language Understanding},
author={Dan Hendrycks and Collin Burns and Steven Basart and Andy Zou and Mantas Mazeika and Dawn Song and Jacob Steinhardt},
booktitle={International Conference on Learning Representations},
year={2021},
url={https://openreview.net/forum?id=d7KBjmI3GmQ}
}

@article{wang2024mmlu,
  title={Mmlu-pro: A more robust and challenging multi-task language understanding benchmark},
  author={Wang, Yubo and Ma, Xueguang and Zhang, Ge and Ni, Yuansheng and Chandra, Abhranil and Guo, Shiguang and Ren, Weiming and Arulraj, Aaran and He, Xuan and Jiang, Ziyan and others},
  journal={arXiv preprint arXiv:2406.01574},
  year={2024}
}

@article{suzgun2022challenging,
  title={Challenging big-bench tasks and whether chain-of-thought can solve them},
  author={Suzgun, Mirac and Scales, Nathan and Sch{\"a}rli, Nathanael and Gehrmann, Sebastian and Tay, Yi and Chung, Hyung Won and Chowdhery, Aakanksha and Le, Quoc V and Chi, Ed H and Zhou, Denny and others},
  journal={arXiv preprint arXiv:2210.09261},
  year={2022}
}

@article{cobbe2021gsm8k,
  title={Training Verifiers to Solve Math Word Problems},
  author={Cobbe, Karl and Kosaraju, Vineet and Bavarian, Mohammad and Chen, Mark and Jun, Heewoo and Kaiser, Lukasz and Plappert, Matthias and Tworek, Jerry and Hilton, Jacob and Nakano, Reiichiro and Hesse, Christopher and Schulman, John},
  journal={arXiv preprint arXiv:2110.14168},
  year={2021}
}

@article{sakaguchi2021winogrande,
  title={Winogrande: An adversarial winograd schema challenge at scale},
  author={Sakaguchi, Keisuke and Bras, Ronan Le and Bhagavatula, Chandra and Choi, Yejin},
  journal={Communications of the ACM},
  volume={64},
  number={9},
  pages={99--106},
  year={2021},
  publisher={ACM New York, NY, USA}
}

@article{allenai:arc,
      author    = {Peter Clark  and Isaac Cowhey and Oren Etzioni and Tushar Khot and
                    Ashish Sabharwal and Carissa Schoenick and Oyvind Tafjord},
      title     = {Think you have Solved Question Answering? Try ARC, the AI2 Reasoning Challenge},
      journal   = {arXiv:1803.05457v1},
      year      = {2018},
}

@article{sap2019socialiqa,
  title={Socialiqa: Commonsense reasoning about social interactions},
  author={Sap, Maarten and Rashkin, Hannah and Chen, Derek and LeBras, Ronan and Choi, Yejin},
  journal={arXiv preprint arXiv:1904.09728},
  year={2019}
}

@inproceedings{clark2019boolq,
    title = "{B}ool{Q}: Exploring the Surprising Difficulty of Natural Yes/No Questions",
    author = "Clark, Christopher  and
      Lee, Kenton  and
      Chang, Ming-Wei  and
      Kwiatkowski, Tom  and
      Collins, Michael  and
      Toutanova, Kristina",
    editor = "Burstein, Jill  and
      Doran, Christy  and
      Solorio, Thamar",
    booktitle = "Proceedings of the 2019 Conference of the North {A}merican Chapter of the Association for Computational Linguistics: Human Language Technologies, Volume 1 (Long and Short Papers)",
    month = jun,
    year = "2019",
    address = "Minneapolis, Minnesota",
    publisher = "Association for Computational Linguistics",
    url = "https://aclanthology.org/N19-1300/",
    doi = "10.18653/v1/N19-1300",
    pages = "2924--2936",
}

@article{kwiatkowski-etal-2019-natural,
    title = "Natural Questions: A Benchmark for Question Answering Research",
    author = "Kwiatkowski, Tom  and
      Palomaki, Jennimaria  and
      Redfield, Olivia  and
      Collins, Michael  and
      Parikh, Ankur  and
      Alberti, Chris  and
      Epstein, Danielle  and
      Polosukhin, Illia  and
      Devlin, Jacob  and
      Lee, Kenton  and
      Toutanova, Kristina  and
      Jones, Llion  and
      Kelcey, Matthew  and
      Chang, Ming-Wei  and
      Dai, Andrew M.  and
      Uszkoreit, Jakob  and
      Le, Quoc  and
      Petrov, Slav",
    editor = "Lee, Lillian  and
      Johnson, Mark  and
      Roark, Brian  and
      Nenkova, Ani",
    journal = "Transactions of the Association for Computational Linguistics",
    volume = "7",
    year = "2019",
    address = "Cambridge, MA",
    publisher = "MIT Press",
    url = "https://aclanthology.org/Q19-1026/",
    doi = "10.1162/tacl_a_00276",
    pages = "452--466",
}

@article{austin2021program,
  title={Program synthesis with large language models},
  author={Austin, Jacob and Odena, Augustus and Nye, Maxwell and Bosma, Maarten and Michalewski, Henryk and Dohan, David and Jiang, Ellen and Cai, Carrie and Terry, Michael and Le, Quoc and others},
  journal={arXiv preprint arXiv:2108.07732},
  year={2021}
}

@article{shi2022language,
  title={Language models are multilingual chain-of-thought reasoners},
  author={Shi, Freda and Suzgun, Mirac and Freitag, Markus and Wang, Xuezhi and Srivats, Suraj and Vosoughi, Soroush and Chung, Hyung Won and Tay, Yi and Ruder, Sebastian and Zhou, Denny and others},
  journal={arXiv preprint arXiv:2210.03057},
  year={2022}
}

@article{chen2021evaluating,
  title={Evaluating large language models trained on code},
  author={Chen, Mark and Tworek, Jerry and Jun, Heewoo and Yuan, Qiming and Pinto, Henrique Ponde De Oliveira and Kaplan, Jared and Edwards, Harri and Burda, Yuri and Joseph, Nicholas and Brockman, Greg and others},
  journal={arXiv preprint arXiv:2107.03374},
  year={2021}
}

@article{rein2023gpqa,
  title={Gpqa: A graduate-level google-proof q\&a benchmark},
  author={Rein, David and Hou, Betty Li and Stickland, Asa Cooper and Petty, Jackson and Pang, Richard Yuanzhe and Dirani, Julien and Michael, Julian and Bowman, Samuel R},
  journal={arXiv preprint arXiv:2311.12022},
  year={2023}
}

@article{dua2019drop,
  title={DROP: A reading comprehension benchmark requiring discrete reasoning over paragraphs},
  author={Dua, Dheeru and Wang, Yizhong and Dasigi, Pradeep and Stanovsky, Gabriel and Singh, Sameer and Gardner, Matt},
  journal={arXiv preprint arXiv:1903.00161},
  year={2019}
}

@article{zhong2023agieval,
  title={Agieval: A human-centric benchmark for evaluating foundation models},
  author={Zhong, Wanjun and Cui, Ruixiang and Guo, Yiduo and Liang, Yaobo and Lu, Shuai and Wang, Yanlin and Saied, Amin and Chen, Weizhu and Duan, Nan},
  journal={arXiv preprint arXiv:2304.06364},
  year={2023}
}

@inproceedings{zellers2019hellaswag,
    title={HellaSwag: Can a Machine Really Finish Your Sentence?},
    author={Zellers, Rowan and Holtzman, Ari and Bisk, Yonatan and Farhadi, Ali and Choi, Yejin},
    booktitle ={Proceedings of the 57th Annual Meeting of the Association for Computational Linguistics},
    year={2019}
}

@article{2017arXivtriviaqa,
       author = {{Joshi}, Mandar and {Choi}, Eunsol and {Weld},
                 Daniel and {Zettlemoyer}, Luke},
        title = "{triviaqa: A Large Scale Distantly Supervised Challenge Dataset for Reading Comprehension}",
      journal = {arXiv e-prints},
         year = 2017,
          eid = {arXiv:1705.03551},
        pages = {arXiv:1705.03551},
archivePrefix = {arXiv},
       eprint = {1705.03551},
}

@inproceedings{bisk2020piqa,
  title={Piqa: Reasoning about physical commonsense in natural language},
  author={Bisk, Yonatan and Zellers, Rowan and Gao, Jianfeng and Choi, Yejin and others},
  booktitle={Proceedings of the AAAI conference on artificial intelligence},
  volume={34},
  pages={7432--7439},
  year={2020}
}

@article{chen2015microsoft,
  title={Microsoft coco captions: Data collection and evaluation server},
  author={Chen, Xinlei and Fang, Hao and Lin, Tsung-Yi and Vedantam, Ramakrishna and Gupta, Saurabh and Doll{\'a}r, Piotr and Zitnick, C Lawrence},
  journal={arXiv preprint arXiv:1504.00325},
  year={2015}
}

@inproceedings{mathew2021docvqa,
  title={Docvqa: A dataset for vqa on document images},
  author={Mathew, Minesh and Karatzas, Dimosthenis and Jawahar, CV},
  booktitle={Proceedings of the IEEE/CVF winter conference on applications of computer vision},
  pages={2200--2209},
  year={2021}
}

@inproceedings{mathew2022infographicvqa,
  title={Infographicvqa},
  author={Mathew, Minesh and Bagal, Viraj and Tito, Rub{\`e}n and Karatzas, Dimosthenis and Valveny, Ernest and Jawahar, CV},
  booktitle={Proceedings of the IEEE/CVF Winter Conference on Applications of Computer Vision},
  pages={1697--1706},
  year={2022}
}

@inproceedings{yue2024mmmu,
  title={Mmmu: A massive multi-discipline multimodal understanding and reasoning benchmark for expert agi},
  author={Yue, Xiang and Ni, Yuansheng and Zhang, Kai and Zheng, Tianyu and Liu, Ruoqi and Zhang, Ge and Stevens, Samuel and Jiang, Dongfu and Ren, Weiming and Sun, Yuxuan and others},
  booktitle={Proceedings of the IEEE/CVF Conference on Computer Vision and Pattern Recognition},
  pages={9556--9567},
  year={2024}
}

@inproceedings{singh2019towards,
  title={Towards vqa models that can read},
  author={Singh, Amanpreet and Natarajan, Vivek and Shah, Meet and Jiang, Yu and Chen, Xinlei and Batra, Dhruv and Parikh, Devi and Rohrbach, Marcus},
  booktitle={Proceedings of the IEEE/CVF conference on computer vision and pattern recognition},
  pages={8317--8326},
  year={2019}
}

@misc{realworldqa,
  title = {RealWorldQA},
  author = {xAI},
  howpublished = {\url{https://x.ai/news/grok-1.5v}},
  year = {2024},
}

@inproceedings{kembhavi2016diagram,
  title={A diagram is worth a dozen images},
  author={Kembhavi, Aniruddha and Salvato, Mike and Kolve, Eric and Seo, Minjoon and Hajishirzi, Hannaneh and Farhadi, Ali},
  booktitle={European conference on computer vision},
  pages={235--251},
  year={2016},
  organization={Springer}
}

@article{masry2022chartqa,
  title={Chartqa: A benchmark for question answering about charts with visual and logical reasoning},
  author={Masry, Ahmed and Long, Do Xuan and Tan, Jia Qing and Joty, Shafiq and Hoque, Enamul},
  journal={arXiv preprint arXiv:2203.10244},
  year={2022}
}

@inproceedings{goyal2017making,
  title={Making the v in vqa matter: Elevating the role of image understanding in visual question answering},
  author={Goyal, Yash and Khot, Tejas and Summers-Stay, Douglas and Batra, Dhruv and Parikh, Devi},
  booktitle={Proceedings of the IEEE conference on computer vision and pattern recognition},
  pages={6904--6913},
  year={2017}
}

@inproceedings{acharya2019tallyqa,
  title={Tallyqa: Answering complex counting questions},
  author={Acharya, Manoj and Kafle, Kushal and Kanan, Christopher},
  booktitle={Proceedings of the AAAI conference on artificial intelligence},
  volume={33},
  pages={8076--8084},
  year={2019}
}

@inproceedings{yang2019spatialsense,
  title={Spatialsense: An adversarially crowdsourced benchmark for spatial relation recognition},
  author={Yang, Kaiyu and Russakovsky, Olga and Deng, Jia},
  booktitle={Proceedings of the IEEE/CVF International Conference on Computer Vision},
  pages={2051--2060},
  year={2019}
}

@article{singh2024global,
  title={Global mmlu: Understanding and addressing cultural and linguistic biases in multilingual evaluation},
  author={Singh, Shivalika and Romanou, Angelika and Fourrier, Cl{\'e}mentine and Adelani, David I and Ngui, Jian Gang and Vila-Suero, Daniel and Limkonchotiwat, Peerat and Marchisio, Kelly and Leong, Wei Qi and Susanto, Yosephine and others},
  journal={arXiv preprint arXiv:2412.03304},
  year={2024}
}

@inproceedings{deutsch-etal-2025-wmt24,
    title = "{WMT}24++: Expanding the Language Coverage of {WMT}24 to 55 Languages {\&} Dialects",
    author = "Deutsch, Daniel  and
      Briakou, Eleftheria  and
      Caswell, Isaac Rayburn  and
      Finkelstein, Mara  and
      Galor, Rebecca  and
      Juraska, Juraj  and
      Kovacs, Geza  and
      Lui, Alison  and
      Rei, Ricardo  and
      Riesa, Jason  and
      Rijhwani, Shruti  and
      Riley, Parker  and
      Salesky, Elizabeth  and
      Trabelsi, Firas  and
      Winkler, Stephanie  and
      Zhang, Biao  and
      Freitag, Markus",
    editor = "Che, Wanxiang  and
      Nabende, Joyce  and
      Shutova, Ekaterina  and
      Pilehvar, Mohammad Taher",
    booktitle = "Findings of the Association for Computational Linguistics: ACL 2025",
    month = jul,
    year = "2025",
    address = "Vienna, Austria",
    publisher = "Association for Computational Linguistics",
    url = "https://aclanthology.org/2025.findings-acl.634/",
    doi = "10.18653/v1/2025.findings-acl.634",
    pages = "12257--12284",
    ISBN = "979-8-89176-256-5",
}

@article{goyal2022flores,
  title={The flores-101 evaluation benchmark for low-resource and multilingual machine translation},
  author={Goyal, Naman and Gao, Cynthia and Chaudhary, Vishrav and Chen, Peng-Jen and Wenzek, Guillaume and Ju, Da and Krishnan, Sanjana and Ranzato, Marc’Aurelio and Guzm{\'a}n, Francisco and Fan, Angela},
  journal={Transactions of the Association for Computational Linguistics},
  volume={10},
  pages={522--538},
  year={2022},
  publisher={MIT Press One Broadway, 12th Floor, Cambridge, Massachusetts 02142, USA~…}
}

@article{artetxe2019cross,
  title={On the cross-lingual transferability of monolingual representations},
  author={Artetxe, Mikel and Ruder, Sebastian and Yogatama, Dani},
  journal={arXiv preprint arXiv:1910.11856},
  year={2019}
}

@article{hsieh2024ruler,
  title={RULER: What's the Real Context Size of Your Long-Context Language Models?},
  author={Hsieh, Cheng-Ping and Sun, Simeng and Kriman, Samuel and Acharya, Shantanu and Rekesh, Dima and Jia, Fei and Zhang, Yang and Ginsburg, Boris},
  journal={arXiv preprint arXiv:2404.06654},
  year={2024}
}

@article{vodrahalli2024michelangelo,
  title={Michelangelo: Long context evaluations beyond haystacks via latent structure queries},
  author={Vodrahalli, Kiran and Ontanon, Santiago and Tripuraneni, Nilesh and Xu, Kelvin and Jain, Sanil and Shivanna, Rakesh and Hui, Jeffrey and Dikkala, Nishanth and Kazemi, Mehran and Fatemi, Bahare and others},
  journal={arXiv preprint arXiv:2409.12640},
  year={2024}
}

@article{kazemi2025big,
  title={Big-bench extra hard},
  author={Kazemi, Mehran and Fatemi, Bahare and Bansal, Hritik and Palowitch, John and Anastasiou, Chrysovalantis and Mehta, Sanket Vaibhav and Jain, Lalit K and Aglietti, Virginia and Jindal, Disha and Chen, Peter and others},
  journal={arXiv preprint arXiv:2502.19187},
  year={2025}
}

@article{zhou2019understanding,
  title={Understanding knowledge distillation in non-autoregressive machine translation},
  author={Zhou, Chunting and Neubig, Graham and Gu, Jiatao},
  journal={arXiv preprint arXiv:1911.02727},
  year={2019}
}

@article{chen2023extending,
  title={Extending context window of large language models via positional interpolation},
  author={Chen, Shouyuan and Wong, Sherman and Chen, Liangjian and Tian, Yuandong},
  journal={arXiv preprint arXiv:2306.15595},
  year={2023}
}

@article{xue2022byt5,
  title={ByT5: Towards a token-free future with pre-trained byte-to-byte models},
  author={Xue, Linting and Barua, Aditya and Constant, Noah and Al-Rfou, Rami and Narang, Sharan and Kale, Mihir and Roberts, Adam and Raffel, Colin},
  journal={Transactions of the Association for Computational Linguistics},
  volume={10},
  pages={291--306},
  year={2022},
  publisher={MIT Press One Broadway, 12th Floor, Cambridge, Massachusetts 02142, USA~…}
}

@inproceedings{xue2021mt5,
  title={mT5: A massively multilingual pre-trained text-to-text transformer},
  author={Xue, Linting and Constant, Noah and Roberts, Adam and Kale, Mihir and Al-Rfou, Rami and Siddhant, Aditya and Barua, Aditya and Raffel, Colin},
  booktitle={Proceedings of the 2021 conference of the North American chapter of the association for computational linguistics: Human language technologies},
  pages={483--498},
  year={2021}
}

@inproceedings{ao2022speecht5,
  title={Speecht5: Unified-modal encoder-decoder pre-training for spoken language processing},
  author={Ao, Junyi and Wang, Rui and Zhou, Long and Wang, Chengyi and Ren, Shuo and Wu, Yu and Liu, Shujie and Ko, Tom and Li, Qing and Zhang, Yu and others},
  booktitle={Proceedings of the 60th annual meeting of the association for computational linguistics (volume 1: Long papers)},
  pages={5723--5738},
  year={2022}
}

@inproceedings{wortsman2022model,
  title={Model soups: averaging weights of multiple fine-tuned models improves accuracy without increasing inference time},
  author={Wortsman, Mitchell and Ilharco, Gabriel and Gadre, Samir Ya and Roelofs, Rebecca and Gontijo-Lopes, Raphael and Morcos, Ari S and Namkoong, Hongseok and Farhadi, Ali and Carmon, Yair and Kornblith, Simon and others},
  booktitle={International conference on machine learning},
  pages={23965--23998},
  year={2022},
  organization={PMLR}
}

@article{vera2025embeddinggemma,
  title={EmbeddingGemma: Powerful and Lightweight Text Representations},
  author={Vera, Henrique Schechter and Dua, Sahil and Zhang, Biao and Naim, Iftekhar and Chen, Feiyang and Cameron, Glenn and Ballantyne, Ian and Black, Kat and Li, Zhe and others},
  journal={arXiv preprint arXiv:2509.20354},
  year={2025},
  url={https://arxiv.org/abs/2509.20354}
}

@inproceedings{press-wolf-2017-using,
    title = "Using the Output Embedding to Improve Language Models",
    author = "Press, Ofir  and
      Wolf, Lior",
    editor = "Lapata, Mirella  and
      Blunsom, Phil  and
      Koller, Alexander",
    booktitle = "Proceedings of the 15th Conference of the {E}uropean Chapter of the Association for Computational Linguistics: Volume 2, Short Papers",
    month = apr,
    year = "2017",
    address = "Valencia, Spain",
    publisher = "Association for Computational Linguistics",
    url = "https://aclanthology.org/E17-2025/",
    pages = "157--163",
}

\end{document}